\documentclass[conference]{IEEEtran}
\IEEEoverridecommandlockouts
\usepackage{cite}
\usepackage{amsmath}
\usepackage{amsfonts}
\usepackage{amssymb}
\usepackage{hyperref}
\usepackage{algorithmic}
\usepackage{graphicx}
\usepackage{textcomp}
\usepackage{xcolor,epsfig,graphicx,booktabs,multirow,float,array,mathrsfs,bm}
\def\BibTeX{{\rm B\kern-.05em{\sc i\kern-.025em b}\kern-.08em
    T\kern-.1667em\lower.7ex\hbox{E}\kern-.125emX}}
\begin{document}

\title{Adaptive Multi-Teacher Knowledge Distillation with Meta-Learning
}

\author{
    \IEEEauthorblockN{Hailin Zhang}
    \IEEEauthorblockA{\textit{Zhejiang University}\\
    zzzhl@zju.edu.cn}
    \and
    \IEEEauthorblockN{Defang Chen}
    \IEEEauthorblockA{
    \textit{Zhejiang University}\\
    defchern@zju.edu.cn}
    \and
    \IEEEauthorblockN{Can Wang}
    \IEEEauthorblockA{
    \textit{Zhejiang University}\\
    wcan@zju.edu.cn}
}

\maketitle

\begin{abstract}
Multi-Teacher knowledge distillation provides students with additional supervision from multiple pre-trained teachers with diverse information sources. Most existing methods explore different weighting strategies to obtain a powerful ensemble teacher, while ignoring the student with poor learning ability may not benefit from such specialized integrated knowledge. To address this problem, we propose Adaptive \textbf{M}ulti-teacher \textbf{K}nowledge \textbf{D}istillation with \textbf{M}eta-Learning (MMKD) to supervise student with appropriate knowledge from a tailored ensemble teacher. With the help of a meta-weight network, the diverse yet compatible teacher knowledge in the output layer and intermediate layers is jointly leveraged to enhance the student performance. Extensive experiments on multiple benchmark datasets validate the effectiveness and flexibility of our methods. Code is available: \href{https://github.com/Rorozhl/MMKD}{https://github.com/Rorozhl/MMKD}.
\end{abstract}


\section{Introduction}
\label{sec:intro}

In recent years, knowledge distillation (KD) has received extensive attention as an effective model compression technology \cite{hinton2015distilling}. It distills knowledge from a large teacher network to a small student network, promoting the student to achieve a prediction similar to the teacher while maintaining less parameters. Existing KD methods generally extract rich forms of teacher knowledge, such as logits \cite{hinton2015distilling} and features \cite{romero2015fitnet,zagoruyko2017paying,chen2021cross} to improve student performance. The teacher-student learning paradigm shows great potential in boosting the student performance with \textit{a single} teacher. We thus expect to gather the wisdom of \textit{multiple teachers} to further bring additional and diverse knowledge to the student training \cite{fukuda2017efficient,chen2020online}.

\begin{figure}
    \begin{minipage}{0.4\linewidth}
        \centering
        \resizebox{1.0\textwidth}{!}{
        \begin{tabular}{c|c} 
            \toprule
            &Accuracy (\%)\\
            \midrule
            Teacher 1 &77.92  \\
            Teacher 2 &78.22 \\
            Teacher 3 &78.39 \\
            Student &64.63 \\
            \bottomrule
        \end{tabular}}%
        \vspace{1em}
    \end{minipage}
    \hspace{0.1em}
    \begin{minipage}{0.53\linewidth}
    \centering
    \includegraphics[width=1.0\textwidth]{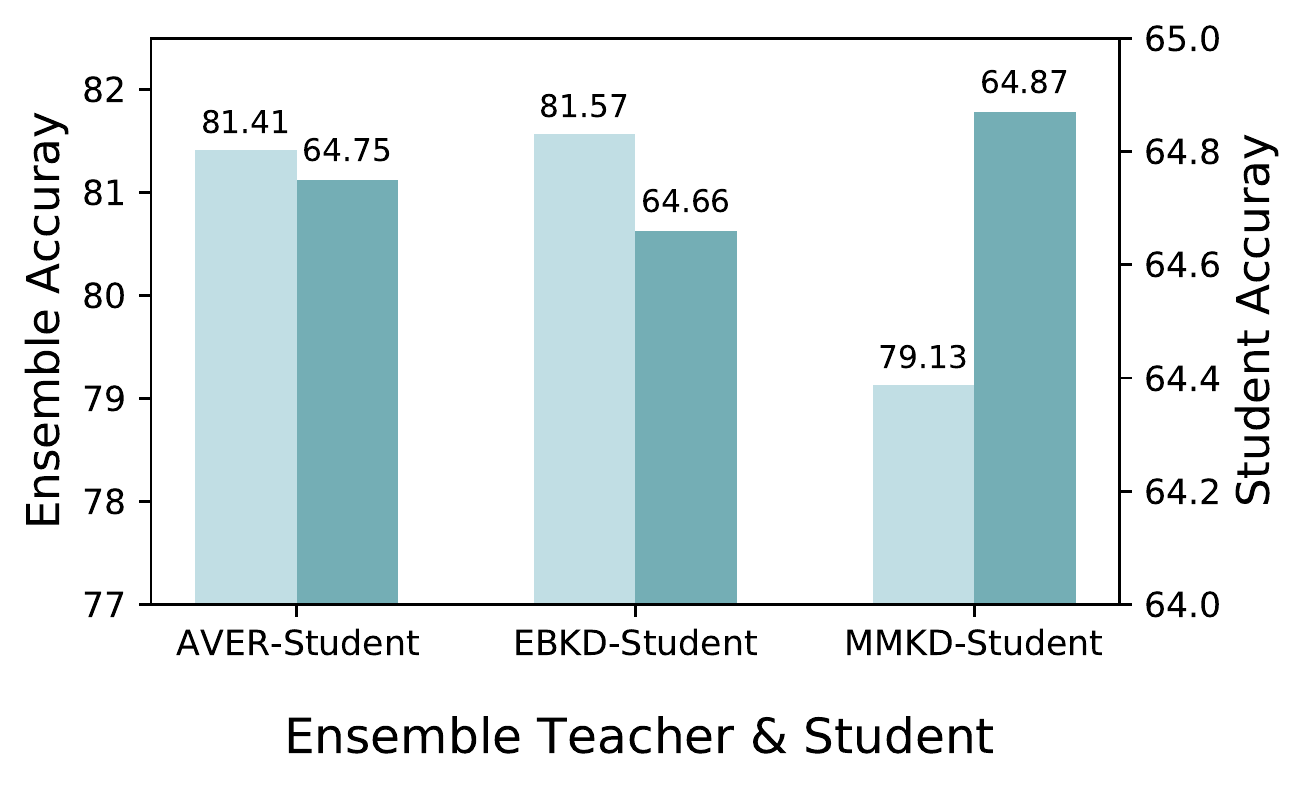}
    \end{minipage}
    \caption{Comparison of model performance with different integration strategies. The accuracies of three teachers and the baseline student are provided in the left Table, while the accuracies of the ensemble teacher and the distilled student are provided in the right Figure.}
    \label{fig:motivation}
\end{figure}

The current multi-teacher KD methods focus on integrating into a \textit{stronger} teacher, which neglects the \textit{knowledge compatibility} between the ensemble teacher and the student, and thus may fail to maximize the benefits of diverse teacher knowledge. 
In fact, knowledge compatibility reflects whether the rich teacher knowledge can be reasonably absorbed by the weak student and we make an empirical study to support our claim. We train three ResNet32x4 \cite{he2016deep} as teachers (with a randomly picked checkpoint) and a MobileNetV2 \cite{sandler2018mobilenetv2} as the student with only labels on the CIFAR-100 dataset. Subsequently, the teachers are integrated with the popular multi-teacher approaches AVER \cite{fukuda2017efficient} and EBKD \cite{kwon2020adaptive}. 
As shown in Figure \ref{fig:motivation}, we find that the teacher obtained by EBKD has a better predictive ability than AVER, but the associated student accuracy is inferior to the AVER. This shows that pursuing a powerful ensemble teacher without considering the knowledge compatibility may not actually bring effective guidance for the student due to its weak receptivity. Therefore, a method that adaptively coordinates the compatibility of knowledge between the ensemble teacher and the student is required for better student training.

To this end, we propose Adaptive \textbf{M}ulti-teacher \textbf{K}nowledge \textbf{D}istillation with \textbf{M}eta-Learning (MMKD), which adopts meta-weight network to integrate the logits and intermediate features of multiple teachers to guide the student in the instance level. Specifically, the probability distributions and pairwise similarity matrices are fed into the meta-weight network to aggregate teacher knowledge in various forms and provide the student with diverse yet compatible knowledge. Besides, we design a \textit{hard buffer} containing difficult samples for guiding the meta-weight network to learn the knowledge that the student struggles to absorb in the knowledge transfer process.
As shown in Figure \ref{fig:motivation}, our MMKD achieves 0.24\% accuracy improvement and beats the competitors, even though the accuracy of ensemble teacher is not the highest.

In a nutshell, with the help of our meta-weight network, we avoid the fine-grained yet difficult patterns extracted by the complex ensemble teacher, and make full use of multiple teachers to effectively provide the student with diverse knowledge. To verify the effectiveness of our approach, we conduct extensive experiments on multiple benchmark datasets and different teacher-student combinations. 

\section{RELATED WORK}

\textbf{Multi-Teacher Knowledge Distillation.} 
Multi-Teacher KD usually combine the soft labels from multiple pre-trained teacher networks to guide the student. Among them, AVER allocates an equal weight to every teachers \cite{fukuda2017efficient}, while RLKD introduces reinforcement learning to filter out the inappropriate teachers before average integration \cite{yuan2021reinforced}. To further exploit the diverse strengths of various teachers, EBKD utilizes information entropy to assign different weights to each teacher \cite{kwon2020adaptive}. In a similar way, CA-MKD quantifies the confidence about teacher predictions through the cross entropy with ground-truth labels \cite{zhang2022confidence}. AEKD proposes a new perspective on the gradient space to examine the diversity of multiple teachers \cite{du2020agree}. These approaches neglect the compatibility between the ensemble teacher and student, while we take it into account.

\textbf{Meta-Learning.} 
Meta-learning aims to make the model learn how to learn such that it can be quickly adapted to new tasks. The famous model-agnostic meta-learning learns parameter initialization with second-order optimization to provide excellent prior knowledge for new tasks \cite{finn2017model}. This idea was gradually adopted in other fields, such as transfer learning scenario \cite{jang2019learning}. The inner loop updates the parameters by minimizing the loss on the target domain model, and the outer loop optimizes the intensity of the source domain model being transferred \cite{jang2019learning}. 
In this paper, we utilize a similar second-order optimization to train our meta-weight network.

\begin{figure*}[t]
    \begin{minipage}[b]{1.0\linewidth}
    \centering
    \includegraphics[width =0.8\textwidth]{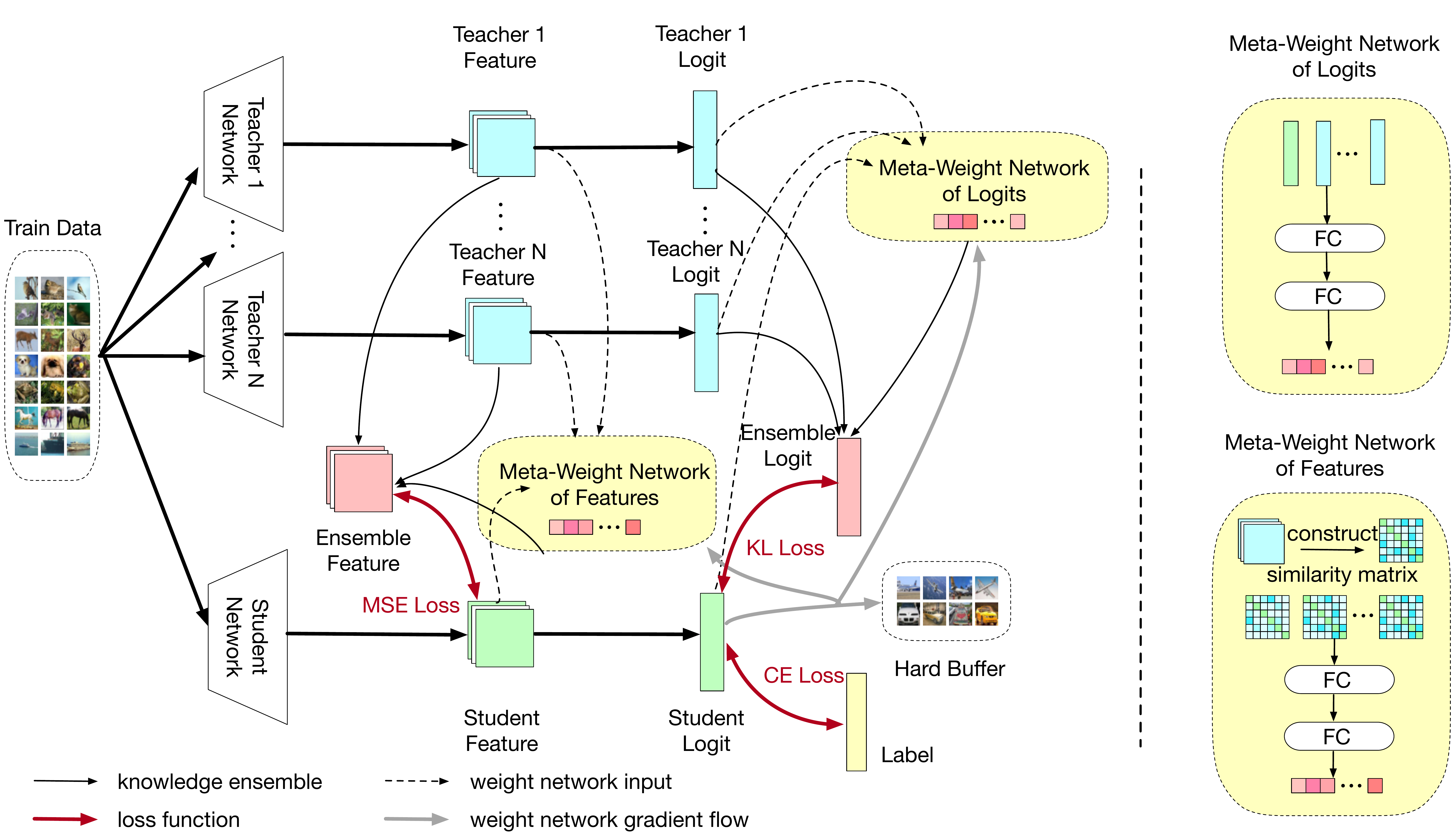}
    \end{minipage}
    \caption{An overview of our MMKD. The integration process for multiple teachers and the gradient backward direction of the meta-weight network are depicted as solid black lines and gray black lines, respectively.}
    \label{fig:structure}
\end{figure*}

\section{METHODOLOGY}

We first briefly introduce necessary notations in our method. Given a dataset $\mathcal{X} = \{(\bm x _1, \bm y_1),(\bm x _2, \bm y_2), \cdots, (\bm x _N, \bm y_N)\}$ with $C$ categories and $N$ samples, which is randomly divided into several mini-batch $\mathcal{B}$ during training. There are $K$ teachers and one student with parameters $\theta^s$. $F \in \mathbb{R}^{b\times h\times w\times c}$ is the last feature maps and $\bm z=[z_1, z_2, \cdots, z_C]$ is the logits vector. We denote $\sigma (\bm z|\tau) = \frac{\exp (z_i / \tau)}{\sum_{j=1}\exp (z_j / \tau)}$ as the smoothed probability distribution with the temperature $\tau$.

\subsection{The Structure of Meta-Weight Network}

In this section, we present the structure of the meta-weight network in the logits and intermediate layers.
\subsubsection{Meta-Weight Network of Logits}
To help the meta-weight network make correct decisions, we input it as much information as possible.
For the logits, we concatenate the output of student and all teachers as:
\begin{eqnarray}
  Z = concat[\bm z^{s}, \bm z^{1}, \bm z^{2}, \cdots, \bm z^{k}], \quad Z\in \mathbb{R}^{C \times (K+1)}.
\end{eqnarray}
It is then projected into a $K$-dimensional weight vector by two fully connected layers $W_r^{1}$ and $W_{r}^{2}$ with $\mathrm{ReLU}(\cdot)$ activation and $softmax(\cdot)$ normalization function, respectively, 
\begin{eqnarray}
  w_{r} = softmax(W_r^{2}  \mathrm{ReLU}(W_r^{1} Z)).
\end{eqnarray}

\subsubsection{Meta-Weight Network of Features}

In the design of the meta-weight network of features, we concatenate the pairwise activation similarity matrices \cite{tung2019similarity,chen2021cross} of the student and teachers as input. To save training time, we only use the feature maps at the penultimate layer to construct similarity matrices
\begin{eqnarray}
  G^{k} = Q^{k} \cdot {Q^{k}}^{T} \quad\quad G^{s} = Q^{s} \cdot {Q^{s}}^{T}, 
\end{eqnarray}
where $Q^k, Q^s \in \mathbb{R}^{b\times (c\times h \times w)}$ represent the features of the $k$th teacher and student after dimension reshape, respectively. The $G^k$ and $G^s$ are the $b\times b$ similarity matrices.

Constructing input in this way can make each teacher and student input with the same dimension.
More importantly, the $b \times b$ matrix encodes the semantic similarity information for each sample pair of the teachers and student, which does not change as the rotation of feature space. The activation similarity matrices of multiple teachers and student are then concatenated as the input of the weight network
\begin{eqnarray}
  G = concat[G^{s}, G^{1}, G^{2}, \cdots, G^{k}].
\end{eqnarray}

We calculate the weight $w_{f}$ for feature matching by feeding $G$ into two fully connected layers $W_f^{1}$ and $W_{f}^{2}$
\begin{eqnarray}
  w_{f} = softmax(W_f^{2}  \mathrm{ReLU}(W_f^{1}  G)).
\end{eqnarray}

With the help of the above effective input information, the weight network can better evaluate the diversity of teachers and the student adaptability.

\subsection{The Training of Student Network}

In the training process of the student network, we utilize the ground-truth labels, the output logits and the intermediate feature representations of multiple teachers as additional knowledge to jointly guide the student.

For the output layer, we input the final probability distributions of multiple teachers and student into the meta-weight network. Then the teacher predictions are integrated by the obtained weight vector $w_{r}$. We measure the prediction distance between the ensemble teacher and student with KL divergence 
\begin{eqnarray}
  \mathcal{L}_r=\frac{1}{b}\sum_{(x,y)\in \mathcal{B}}\sum_{k=1}^{K}w^{k}_r\sum_{c=1}^{C}\sigma (\bm z^{k}_c|\tau)\log \frac{\sigma (\bm z^{k}_c|\tau)}{\sigma (\bm z^s_c|\tau)}.
\end{eqnarray}

For the intermediate layer, we obtain the weight vector $w_{f}$ via feeding the pairwise similarity matrices of the penultimate features of multiple teachers and student into the meta-weight network. Since the teacher and student features may be in different spaces, we introduce additional convolution operation $u(\cdot)$ to transform student feature for alignment \cite{chen2022simkd}. We measure the feature similarity through Mean-Square-Error (MSE) loss function. Therefore, the feature information can be aggregated with the calculated weights
\begin{eqnarray}
  \mathcal{L}_f = \frac{1}{b}\sum_{(x,y)\in \mathcal{B}}\sum_{k}^{k}w^{k}_{f}||F^{k}-u(F^{s})||^2_2.
\end{eqnarray}

The labels are used as an effective signal in supervised learning by calculating the cross entropy between labels and student predictions to measure the classification ability
\begin{eqnarray}
  \mathcal{L}_{CE} = -\frac{1}{b}\sum_{(x,y)\in \mathcal{B}}\sum_{c=1}^{C}\bm y_{c}\log(\sigma^{s} (\bm z^s_c|\tau)).
\end{eqnarray}

Overall, we train the student model using three losses in the inner loop for our MMKD, where $\alpha$ and $\beta$ are the hyper-parameters to balance each loss
\begin{eqnarray}
  \mathcal{L}_{total}=\mathcal{L}_{CE}+\alpha \mathcal{L}_{f}+\beta \mathcal{L}_{r}.
  \label{eq:total}
\end{eqnarray}

\subsection{The Training of Meta-Weight Network}

In the meta-learning framework, validation set is usually used in the outer loop to evaluate the effect on the target task. However, multiple benchmark datasets do not have explicit validation set. Separating a validation set from the training set may lead to the insufficient model training and poor convergence due to the data reduction. Inspired by the \textit{experience replay strategy} in reinforcement learning \cite{mnih2013playing}, we maintain a hard buffer $\mathcal H$ containing difficult samples as the validation set to measure the effectiveness for student. 

For simplicity, we define $\phi$ as the total parameters of the meta-weight network, which includes $W_r^{1}$, $W_r^{2}$, $W_f^{1}$ and $W_f^{2}$. We copy the student as $s'$ and update it with $M$ rounds on the difficult samples according to the Equation (\ref{eq:total}), and denote it as $f(\theta^{s'}_M|x \in \mathcal H, \phi)$. We then optimize $\phi$ to minimize the cross-entropy loss on the difficult samples. 
\begin{eqnarray}
\begin{array}{ll}
  \underset{\phi}{\min} & E(\theta^{s'}_M|x \in \mathcal H) \\
  \text {s.t.} & \theta_i^{s'} = O_i(\theta_{i - 1}^{s'}, \phi) \qquad i=1, 2, \cdots, M
\end{array}
\end{eqnarray}
where $E(\theta^{s'}_M|x \in \mathcal H)$ represents the cross-entropy loss of the student copy $s'$, and $O_i(\cdot)$ is the operation of the optimization algorithm such as SGD in step $i$. We adopt Reverse-HG \cite{franceschi2017forward} to calculate the gradient of $\phi$ based on the above equation and update the meta-weight network in the outer loop.

In our strategy, we adopt the idea of second-order optimization to alternatively update the student and meta-weight network such that we can easily find the optimal solution of meta-weight network under the restriction of inner loop, and help student effectively integrate teacher knowledge which takes into account both diversity and compatibility. More importantly, it avoids the student from being adversely affected by the meta-weight network during the training process, and ensures the student to adjust the direction of optimization according to appropriate knowledge.

\begin{table*}[t]
    \renewcommand\arraystretch{1.05}
	\centering
	\caption{Top-1 test accuracy of multi-teacher KD methods on CIFAR-100 (teachers with the same architectures)}
    \resizebox{0.95\textwidth}{!}{
    \begin{tabular}{c|cccccc|c}
        \toprule
        \multirow{2}*{Teacher} &VGG13 &ResNet32x4 &ResNet32x4 &WRN-40-2 &WRN-40-2 &ResNet20x4  &\multirow{3}*{ARI(\%)} \\
        &75.17$\pm$0.18 &79.31$\pm$0.14 &79.31$\pm$0.14 &76.62$\pm$0.26 &76.62$\pm$0.26 &78.632$\pm$0.24\\
        Ensemble &77.07 &81.16 &81.16 &79.62 &79.62  &80.81\\
        \midrule
        \multirow{2}*{Student}&VGG8 &MobileNetV2 &VGG8 &MobileNetV2 &WRN-40-1 &ShuffleNetV1 &\multirow{2}*{/} \\
        &70.74$\pm$0.40 &65.64$\pm$0.19 &70.74$\pm$0.40 &65.64$\pm$0.19 &71.93$\pm$0.22 &71.70$\pm$0.43 \\
        \midrule
        AVER \cite{fukuda2017efficient}  &73.98$\pm$0.13 &68.42$\pm$0.06 &73.23$\pm$0.35 &69.67$\pm$0.01 &74.56$\pm$0.13 &75.73$\pm$0.02 &49.97\%\\
        FitNet-MKD \cite{romero2015fitnet} &74.05$\pm$0.07 &68.46$\pm$0.49 &73.24$\pm$0.24 &69.29$\pm$0.42 &74.95$\pm$0.30 &75.98$\pm$0.06 &46.97\%\\
        EBKD \cite{kwon2020adaptive} &73.97$\pm$0.34 &68.06$\pm$0.01 &73.63$\pm$0.31 &69.17$\pm$0.11 &74.37$\pm$0.23 &75.82$\pm$0.26 &53.64\%\\
        AEKD-logits \cite{du2020agree} &73.82$\pm$0.09 &68.39$\pm$0.13 &73.22$\pm$0.29 &69.56$\pm$0.34 &74.18$\pm$0.25&75.93$\pm$0.32 &54.87\%\\
        AEKD-feature \cite{du2020agree} &73.99$\pm$0.15 &68.18$\pm$0.06 &73.38$\pm$0.16 &69.44$\pm$0.25 &74.96$\pm$0.18 &76.86$\pm$0.03 &43.16\%\\
        CA-MKD \cite{zhang2022confidence} &74.27$\pm$0.16 &69.19$\pm$0.04 &75.08$\pm$0.07 &70.87$\pm$0.14 &75.27$\pm$0.21 &77.19$\pm$0.49 &11.98\%\\
        \midrule
        \bf MMKD &\bf74.86$\pm$0.07 &\bf69.70$\pm$0.04 &\bf75.66$\pm$0.18 &\bf71.23$\pm$0.03 &\bf75.61$\pm$0.11 &\bf77.76$\pm$0.35 &/\\
        \bottomrule
    \end{tabular}}%
    \label{tab:ex-MMKD-1}
\end{table*}


\section{EXPERIMENT}

In this section, we verify the effectiveness and robustness of our MMKD on multiple benchmark datasets, such as CIFAR-100, Stanford Dogs and Tiny-ImageNet, and we utilize multiple groups of teacher-student architectures. 

\textbf{Experiment Settings.} We generally follow the setting of previous works \cite{chen2021cross, wang2023semckd}. The learning rate starts from 0.1 on CIFAR-100, which is divided by 10 at 150th, 180th and 210th of the total 240 epochs. All methods are optimized with SGD, and our meta-weight network utilizes the Adam with a learning rate 1e-3, which is updated every 5 batches. Besides, the temperature is set to 4, the $\alpha$ is set to 1 in all methods and $\beta$ is set to 10 or 100 in our method. We conduct three trials for each experiment setting and report the mean and std. 

\subsection{Distillation Performance}

We integrate three teacher networks with the same structure but different initialization parameters and distillation strategies to guide the student. We use ARI (Average Relative Improvement)  \footnote{$ARI=\frac{1}{M}\sum_{i=1}^{M}\frac{Acc^i_{MMKD} - Acc^i_{BKD}}{Acc^i_{BKD} - Acc^i_{Stu}}\times 100\%$, where $M$ represents the number of teacher-student combinations, $Acc^i_{MMKD}$, $Acc^i_{BKD}$, $Acc^i_{Stu}$ are the accuracy of our MMKD, other comparison 
methods, and the student trained from scratch, respectively.} to measure the performance improvement of our MMKD against the comparison methods \cite{tian2020contrastive,chen2021cross}. 

As shown in Table~\ref{tab:ex-MMKD-1} and~\ref{tab:ex-MMKD-2}, our MMKD surpasses the current popular single-teacher and multi-teacher KD methods in all teacher-student combinations.
Note that even the simplest multi-teacher KD method AVER can achieve better results than most single-teacher KD methods, which verifies the benefit of learning from multiple teachers. 
However, we also find that the multi-teacher KD methods are not always better than the single-teacher KD method. Although multiple teachers have more diverse knowledge, it may still fail to achieve the decent distillation performance without considering the compatibility of knowledge between the ensemble teacher and student in the knowledge transfer process. This is consistent with our pilot study in Figure~\ref{fig:motivation}. 

\begin{table}[t]
    \renewcommand\arraystretch{1.05}
	\centering
	\caption{Top-1 test accuracy of MMKD compared to single-teacher knowledge distillation methods.}
    \resizebox{0.46\textwidth}{!}{
    \begin{tabular}{c|cccc} 
        \toprule
        \multirow{2}*{Teacher}&ResNet32x4 &WRN-40-2  &WRN-40-2  \\
        &79.31$\pm$0.14 &76.62$\pm$0.26 &76.62$\pm$0.26\\
        \midrule
        \multirow{2}*{Student} &MobileNetV2 &MobileNetV2  &WRN-40-1 \\
        &65.64$\pm$0.19 &65.64$\pm$0.19 
        &71.93$\pm$0.22\\
        \midrule
        KD \cite{hinton2015distilling} &67.57$\pm$0.10 &69.31$\pm$0.20 &74.22$\pm$0.09 \\
        FitNet \cite{romero2015fitnet}  &67.87$\pm$0.08 &69.01$\pm$0.18 &74.28$\pm$0.15\\
        AT \cite{zagoruyko2017paying} &67.38$\pm$0.21 &69.18$\pm$0.37 &74.83$\pm$0.15 \\
        VID \cite{ahn2019variational} &67.78$\pm$0.13  &68.57$\pm$0.11 &74.37$\pm$0.22\\
        CRD \cite{tian2020contrastive} &69.04$\pm$0.16 &70.14$\pm$0.06 &74.82$\pm$0.06\\
        SemCKD \cite{chen2021cross} &68.86$\pm$0.26 &69.61$\pm$0.05 & 74.41$\pm$0.16 \\
        SRRL \cite{yang2021knowledge} &68.77$\pm$0.06 &69.44$\pm$0.13 &74.60$\pm$0.04\\
        \midrule
        \bf MMKD &\bf69.70$\pm$0.04 &\bf71.23$\pm$0.03 &\bf75.61$\pm$0.11\\
        \bottomrule
    \end{tabular}}
    \label{tab:ex-MMKD-2}
\end{table}

\begin{table*}[h]
	\renewcommand\arraystretch{1.05}
	\centering
	\caption{Top-1 test accuracy of multi-teacher KD methods on Stanford Dogs and Tiny-ImageNet (teachers with the same architectures)}
	\resizebox{0.65\textwidth}{!}{
		\begin{tabular}{c|cc|cc} 
			\toprule
			Dataset &\multicolumn{2}{c|}{Stanford Dogs}&\multicolumn{2}{c}{Tiny-ImageNet}\\
			\midrule
			\multirow{2}*{Teacher}&ResNet101 &ResNet34x4  &ResNet32x4 &VGG13 \\
			&68.39$\pm$1.44 &66.07$\pm$0.51  &53.38$\pm$0.11 &49.17$\pm$0.33\\
			\midrule
			\multirow{2}*{Student} &ShuffleNetV2x0.5 &ShuffleNetV2x0.5  &MobileNetV2 &MobileNetV2 \\
			&59.36$\pm$0.73 &59.36$\pm$0.73 
			&39.46$\pm$0.38 &39.46$\pm$0.38 \\
			\midrule
			AVER \cite{fukuda2017efficient} &65.13$\pm$0.13 &63.46$\pm$0.21 &41.78$\pm$0.15 &41.87$\pm$0.11 \\
			FitNet-MKD \cite{romero2015fitnet} &65.23$\pm$0.37 &63.71$\pm$0.31 &41.52$\pm$0.21 &41.46$\pm$0.14\\
			EBKD \cite{kwon2020adaptive} &64.28$\pm$0.13 &64.19$\pm$0.11 &41.24$\pm$0.11 &41.46$\pm$0.24\\
			AEKD-logits \cite{du2020agree} &65.18$\pm$0.24 &63.97$\pm$0.14 &41.46$\pm$0.28 &41.19$\pm$0.23 \\
			AEKD-feature \cite{du2020agree} &64.91$\pm$0.21 &62.13$\pm$0.29 &42.03$\pm$0.12 &41.56$\pm$0.14 \\
			CA-MKD \cite{zhang2022confidence} &64.09$\pm$0.35 &64.28$\pm$0.20 &43.90$\pm$0.09 &42.65$\pm$0.05\\
			\midrule
			\bf MMKD &\bf65.46$\pm$0.22 &\bf64.55$\pm$0.17 &\bf44.10$\pm$0.04 &\bf44.15$\pm$0.15 \\
			\bottomrule
	\end{tabular}}%
	\label{tab:ex-MMKD-3}
\end{table*}
In addition, as shown in Table~\ref{tab:ex-MMKD-3}, our MMKD also performs well on the Stanford Dogs and Tiny-ImageNet datasets and obtains 17.23\% and 26.23\% average relative improvement compared with CA-MKD \cite{zhang2022confidence}. This demonstrates the effectiveness of integrating teacher knowledge to guide the student via meta-weight network. Besides, Table~\ref{tab:ex-MMKD-4} and~\ref{tab:ex-MMKD-5} show that our MMKD still outperform competitors even when there are more significant architecture differences among teachers.

\begin{table*}[htbp]
    \renewcommand\arraystretch{1.05}
	\centering
	\caption{Top-1 test accuracy of MKD methods on CIFAR-100 (three teachers with the different architectures)}
    \resizebox{0.8\textwidth}{!}{
    \begin{tabular}{c|cc|cc|cc} 
        \toprule
        \multirow{3}*{Teacher} &ResNet56 &73.47 &ResNet8 &59.32 &VGG11 &71.52\\
        &ResNet20x4 &78.39 &WRN-40-2 &76.51 &VGG13 &75.19\\
        &VGG13 &75.19 &ResNet20x4 &78.39 &ResNet32x4 &79.31\\
        \midrule
        Student &VGG8 &70.74$\pm$0.40 &ResNet8x4 &72.79$\pm$0.14 &VGG8 &70.74$\pm$0.40\\
        \midrule
        AVER \cite{fukuda2017efficient} &\multicolumn{2}{c|}{75.11$\pm$0.57} &\multicolumn{2}{c|}{75.16$\pm$0.11} &\multicolumn{2}{c}{73.59$\pm$0.06}\\
        FitNet-MKD \cite{romero2015fitnet} &\multicolumn{2}{c|}{75.06$\pm$0.13} &\multicolumn{2}{c|}{75.21$\pm$0.12} &\multicolumn{2}{c}{73.43$\pm$0.08}\\
        EBKD \cite{kwon2020adaptive} &\multicolumn{2}{c|}{74.18$\pm$0.22} &\multicolumn{2}{c|}{75.44$\pm$0.29} &\multicolumn{2}{c}{73.40$\pm$0.06}\\
        AEKD-logits \cite{du2020agree} &\multicolumn{2}{c|}{75.17$\pm$0.30} &\multicolumn{2}{c|}{73.93$\pm$0.17} &\multicolumn{2}{c}{73.45$\pm$0.08}\\
        AEKD-feature \cite{du2020agree} &\multicolumn{2}{c|}{74.69$\pm$0.57} &\multicolumn{2}{c|}{73.98$\pm$0.18} &\multicolumn{2}{c}{74.15$\pm$0.08}\\
        CA-MKD \cite{zhang2022confidence} &\multicolumn{2}{c|}{75.53$\pm$0.14} &\multicolumn{2}{c|}{75.27$\pm$0.18} &\multicolumn{2}{c}{74.63$\pm$0.17}\\
        \midrule
        \bf MMKD &\multicolumn{2}{c|}{\bf 76.21$\pm$0.11} &\multicolumn{2}{c|}{\bf 76.29$\pm$0.16} &\multicolumn{2}{c}{\bf 75.78$\pm$0.04}\\
        \bottomrule
    \end{tabular}}%
    \label{tab:ex-MMKD-4}
\end{table*}

\begin{table*}[htbp]
    \renewcommand\arraystretch{1.05}
	\centering
	\caption{Top-1 test accuracy of MKD methods on CIFAR-100 (five teachers with the different architectures)}
    \resizebox{0.8\textwidth}{!}{
    \begin{tabular}{c|cc|cc|cc} 
        \toprule
        \multirow{5}*{Teacher} &ResNet8 &59.32 &VGG11 &71.52 &ResNet8 &59.32 \\
        &VGG11 &71.52 &ResNet56 &73.47 &VGG11 &71.52\\
        &ResNet56 &73.47 &VGG13 &75.19 &VGG13 &75.19\\
        &VGG13 &75.19 &ResNet20x4 &78.39 &WRN-40-2 &76.51\\
        &ResNet32x4 &79.31 &ResNet32x4 &79.31 &ResNet20x4 &78.39\\
        \midrule
        Student &VGG8 &70.74$\pm$0.40 &VGG8 &70.74$\pm$0.40 &MobileNetV2 &65.64$\pm$0.19\\
        \midrule
        AVER \cite{fukuda2017efficient} &\multicolumn{2}{c|}{74.47$\pm$0.47} &\multicolumn{2}{c|}{74.48$\pm$0.12} &\multicolumn{2}{c}{69.41$\pm$0.04}\\
        FitNet-MKD \cite{romero2015fitnet} &\multicolumn{2}{c|}{74.34$\pm$0.20} &\multicolumn{2}{c|}{74.24$\pm$0.22} &\multicolumn{2}{c}{69.61$\pm$0.25}\\
        EBKD \cite{kwon2020adaptive} &\multicolumn{2}{c|}{74.37$\pm$0.07} &\multicolumn{2}{c|}{73.94$\pm$0.29} &\multicolumn{2}{c}{69.26$\pm$0.64}\\
        AEKD-logits \cite{du2020agree} &\multicolumn{2}{c|}{73.53$\pm$0.10} &\multicolumn{2}{c|}{74.90$\pm$0.17} &\multicolumn{2}{c}{69.28$\pm$0.21}\\
        AEKD-feature \cite{du2020agree} &\multicolumn{2}{c|}{74.02$\pm$0.08} &\multicolumn{2}{c|}{75.06$\pm$0.03} &\multicolumn{2}{c}{69.41$\pm$0.21}\\
        CA-MKD \cite{zhang2022confidence} &\multicolumn{2}{c|}{74.64$\pm$0.23} &\multicolumn{2}{c|}{75.02$\pm$0.21} &\multicolumn{2}{c}{70.30$\pm$0.51}\\
        \midrule
        \bf MMKD &\multicolumn{2}{c|}{\bf 75.42$\pm$0.31} &\multicolumn{2}{c|}{\bf 75.78$\pm$0.13} &\multicolumn{2}{c}{\bf 71.30$\pm$0.20}\\
        \bottomrule
    \end{tabular}}%
    \label{tab:ex-MMKD-5}
\end{table*}

\begin{figure}[tbp]
    \centering
    \includegraphics[width =0.48\textwidth]{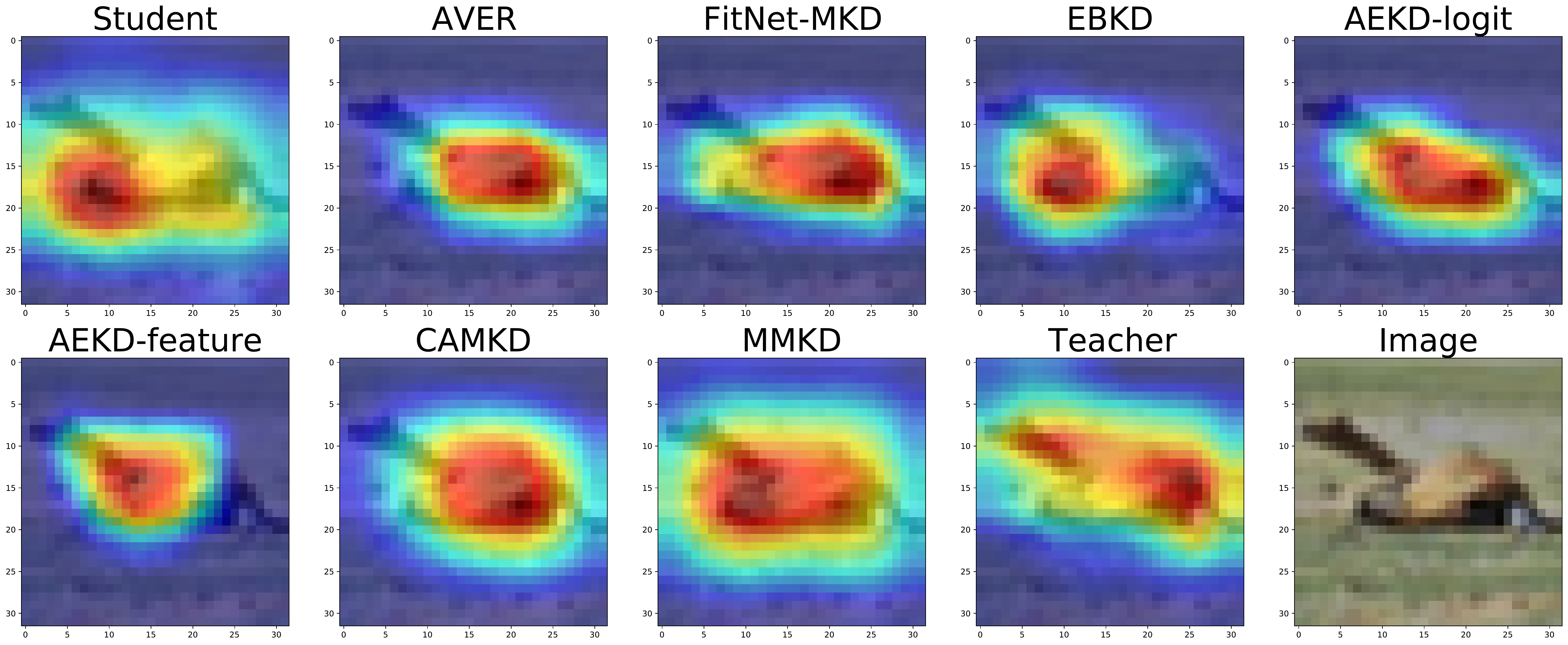}
    \caption{Grad-CAM visualization of multi-teacher knowledge distillation methods for WRN-40-2 \& WRN-40-1 on CIFAR-100.}
    \label{fig:MMKD-grad}
\end{figure}

\begin{figure}[h]
    \centering
    \includegraphics[width =0.96\columnwidth]{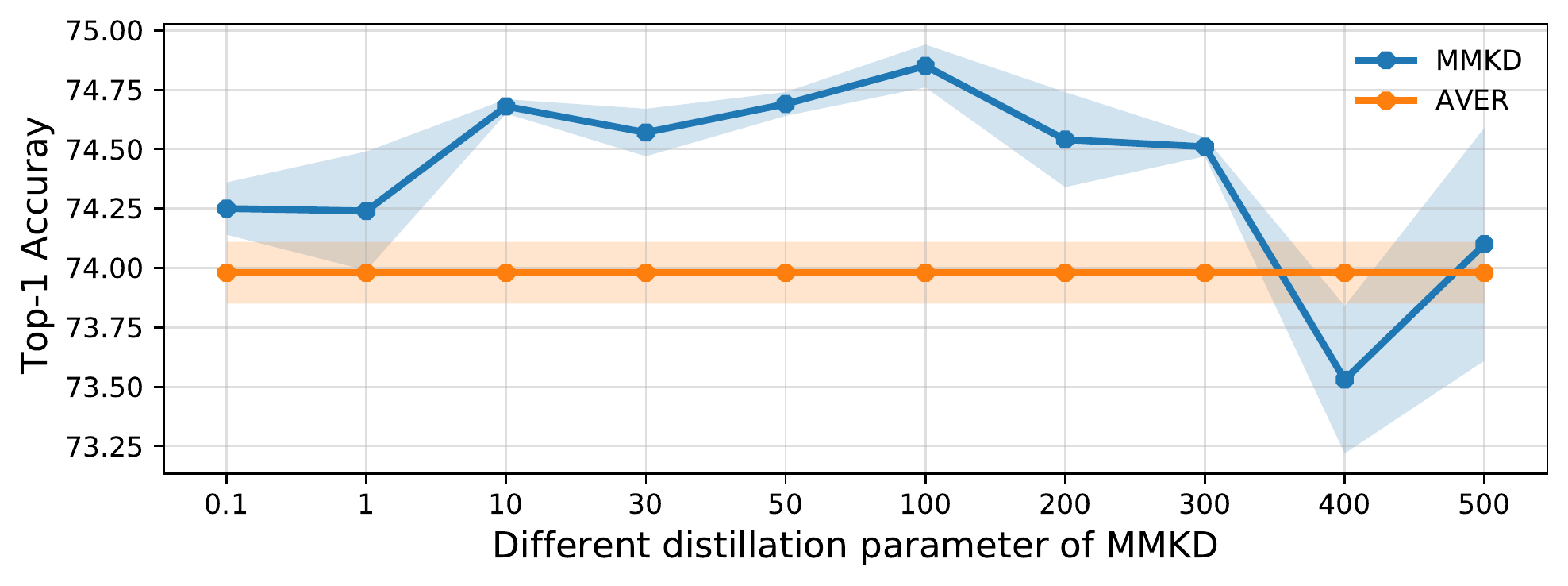}
    \caption{Impact of the hyper-parameter $\beta$ for VGG13 \& VGG8 on CIFAR-100.}
    \label{fig:MMKD-gamma}
\end{figure}

To further intuitively explain the advantages of our MMKD, we use Grad-Cam \cite{selvaraju2017grad} to draw the heat maps of student, distilled student and teacher in Figure \ref{fig:MMKD-grad}. We find that some compared methods pay attention to the background or fail to fully recognize the object, especially for AVER, EBKD and AEKD-logit. We speculate that these methods ignore the feature knowledge and only use the output of teachers, resulting in that the intermediate layers of student are still unable to extract higher-order semantic information. Besides, the highlighted area of MMKD is very similar to teacher, which shows that MMKD can effectively integrate teacher knowledge to improve feature expression ability.

\subsection{Sensitivity Analysis}

We evaluate the impact of hyper-parameter $\beta$ on the performance of distillation, and $\beta$ ranges from 0.1 to 500. Figure \ref{fig:MMKD-gamma} shows that 
as $\beta$ gradually increases, the performance gap between MMKD and AVER gradually enlarges, which indicates that the feature information plays an vital role in the student performance improvement. However, too large $\beta$ makes feature matching dominant in the three losses, which leads to a decline in model accuracy and unstable effect.

\begin{figure}[t]
    \centering
    \includegraphics[width =0.96\columnwidth]{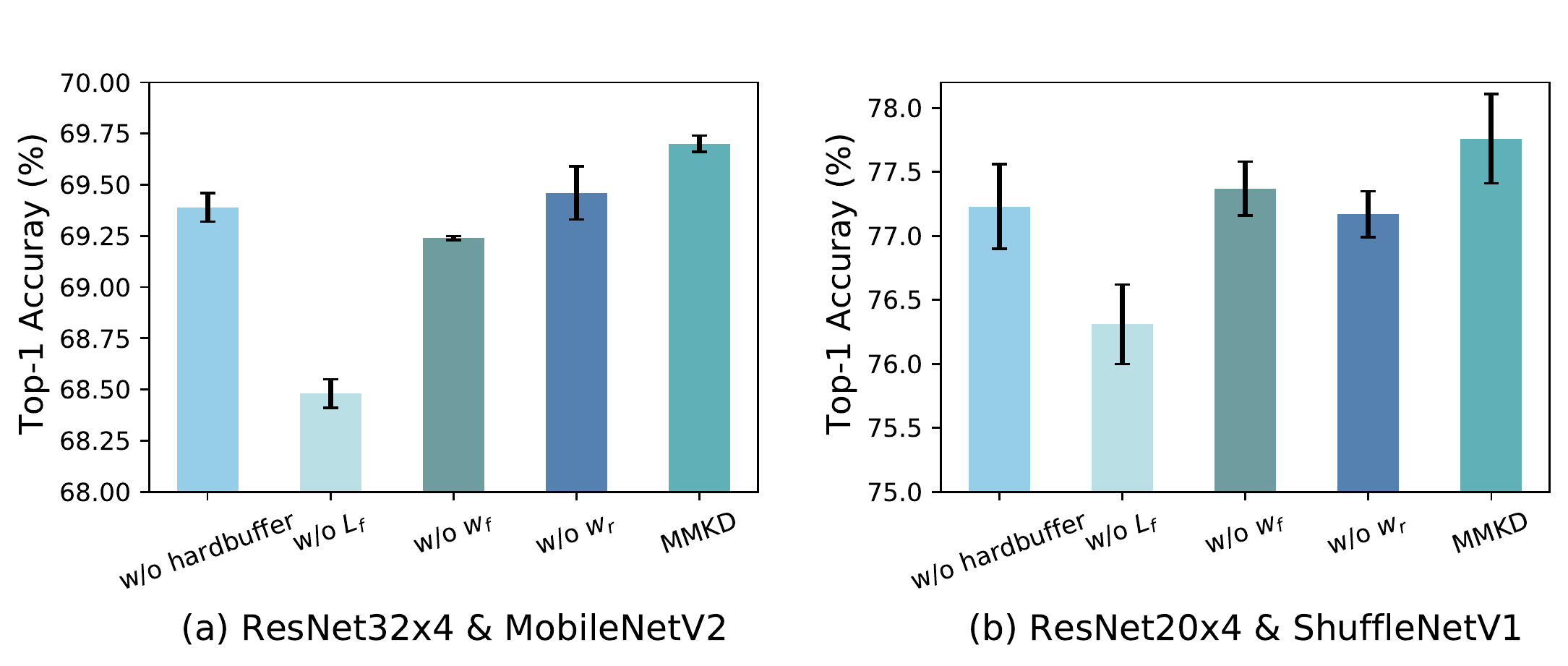}
    \caption{The ablation study on our MMKD.}
    \label{fig:MMKD-ablation}
\end{figure}

\subsection{Ablation Study}

We conduct a series of ablation experiments to verify the effectiveness of key components involved in our MMKD. As shown in Figure \ref{fig:MMKD-ablation}, we observe the accuracy reduction as we remove the hard buffer, which proves that it is necessary to optimize the meta-weight network towards difficult knowledge. Removing the weight of output layer or the intermediate layer causes a significantly loss of accuracy, which means our meta-weight network plays a key role in effectively improving distillation performance. It is worth noting that there is a considerably drop in student accuracy when the feature loss is removed, which further confirms the importance of feature information in the knowledge transfer process.

\section{CONCLUSION}

In this paper, 
we propose a novel approach named MMKD to coordinate the knowledge compatibility of the ensemble teacher and student, which aggregates output predictions and intermediate features of multiple teachers via a meta-weight network. In this way, the student performance can be steadily improved under the guidance of compatible ensemble knowledge. Extensive experimental results validate the superiority of our method. A potential future work is to explore how to apply multi-teacher knowledge distillation techniques in accelerating the sampling speed of the popular diffusion models \cite{chen2023geometric}.


\bibliographystyle{IEEEbib}
\bibliography{refs}

\end{document}